\documentclass[10pt,twocolumn,letterpaper]{article}

\usepackage{cvpr}
\usepackage{times}
\usepackage{epsfig}
\usepackage{graphicx}
\usepackage{amsmath}
\usepackage{amssymb}

\usepackage{array}
\usepackage{epstopdf, booktabs}
\usepackage{enumitem}
\usepackage{subfigure}
\usepackage[english]{babel}
\usepackage{multirow}
\usepackage{lipsum}
\usepackage[linesnumbered,ruled]{algorithm2e}

\DeclareMathAlphabet{\mathpzc}{OT1}{pzc}{m}{it} % \mathcal small letters 

\newcolumntype{P}[1]{>{\centering\arraybackslash}p{#1}}
\newcolumntype{M}[1]{>{\centering\arraybackslash}m{#1}}

\usepackage[pagebackref=true,breaklinks=true,letterpaper=true,colorlinks,bookmarks=false]{hyperref}
\cvprfinalcopy % *** Uncomment this line for the final submission

\def\cvprPaperID{181} % *** Enter the CVPR Paper ID here

\begin{document}

%%%%%%%%% TITLE
%\title{Practical Block-wise Neural Network Architecture Generation}
%
%\author{First Author\\
%Institution1\\
%Institution1 address\\
%{\tt\small firstauthor@i1.org}
%% For a paper whose authors are all at the same institution,
%% omit the following lines up until the closing ``}''.
%% Additional authors and addresses can be added with ``\and'',
%% just like the second author.
%% To save space, use either the email address or home page, not both
%\and
%Second Author\\
%Institution2\\
%First line of institution2 address\\
%{\tt\small secondauthor@i2.org}
%}

\title{Practical Block-wise Neural Network Architecture Generation}

\author{Zhao Zhong\textsuperscript{1,3}\thanks{This work was done when Zhao Zhong worked as an intern at SenseTime Research.}, Junjie Yan\textsuperscript{2},Wei Wu\textsuperscript{2},Jing Shao\textsuperscript{2},Cheng-Lin Liu\textsuperscript{1,3,4}\\
	\emph{\textsuperscript{1}National Laboratory of Pattern Recognition,Institute of Automation, Chinese Academy of Sciences}\\
	\emph{\textsuperscript{2} SenseTime Research} \quad \emph{\textsuperscript{3} University of Chinese Academy of Sciences}\\	
	\emph{\textsuperscript{4} CAS Center for Excellence of Brain Science and Intelligence Technology}\\
	\emph{Email: \{zhao.zhong, liucl\}@nlpr.ia.ac.cn, \{yanjunjie, wuwei, shaojing\}@sensetime.com}}

\maketitle
\thispagestyle{empty}

%%%%%%%%% ABSTRACT
\begin{abstract}
Convolutional neural networks have gained a remarkable success in computer vision.
However, most usable network architectures are hand-crafted and usually require expertise and elaborate design.
In this paper, we provide a block-wise network generation pipeline called BlockQNN which automatically builds high-performance networks using the Q-Learning paradigm with epsilon-greedy exploration strategy.
The optimal network block is constructed by the learning agent which is trained sequentially to choose component layers. We stack the block to construct the whole auto-generated network.
To accelerate the generation process, we also propose a distributed asynchronous framework and an early stop strategy.
The block-wise generation brings unique advantages:
(1) it performs competitive results in comparison to the hand-crafted state-of-the-art networks on image classification, additionally, the best network generated by BlockQNN achieves $3.54\%$ top-$1$ error rate on CIFAR-$10$ which beats all existing auto-generate networks.
(2) in the meanwhile, it offers tremendous reduction of the search space in designing networks which only spends $3$ days with $32$ GPUs,
and (3) moreover, it has strong generalizability that the network built on CIFAR also performs well on a larger-scale ImageNet dataset.
\end{abstract}

%%%%%%%%% BODY TEXT
\section{Introduction}

% ======= Background
During the last decades, Convolutional Neural Networks (CNNs) have shown remarkable potentials almost in every field in the computer vision society~\cite{lecun2015deep}.
For example, thanks to the network evolution from AlexNet~\cite{krizhevsky2012imagenet}, VGG~\cite{simonyan2014very}, Inception~\cite{szegedy2015going} to ResNet~\cite{he2015deep}, the top-5 performance on ImageNet challenge steadily increases from $83.6\%$ to $96.43\%$. 
However, as the performance gain usually requires an increasing network capacity, a high-performance network architecture generally possesses a tremendous number of possible configurations about the number of layers, hyperparameters in each layer and type of each layer.
It is hence infeasible for manually exhaustive search, and the design of successful hand-crafted networks heavily rely on expert knowledge and experience.
Therefore, constructing the network in a smart and automatic manner remains an open problem.

\begin{figure*}[tbp]
	\centering
	\includegraphics[width=1\textwidth]{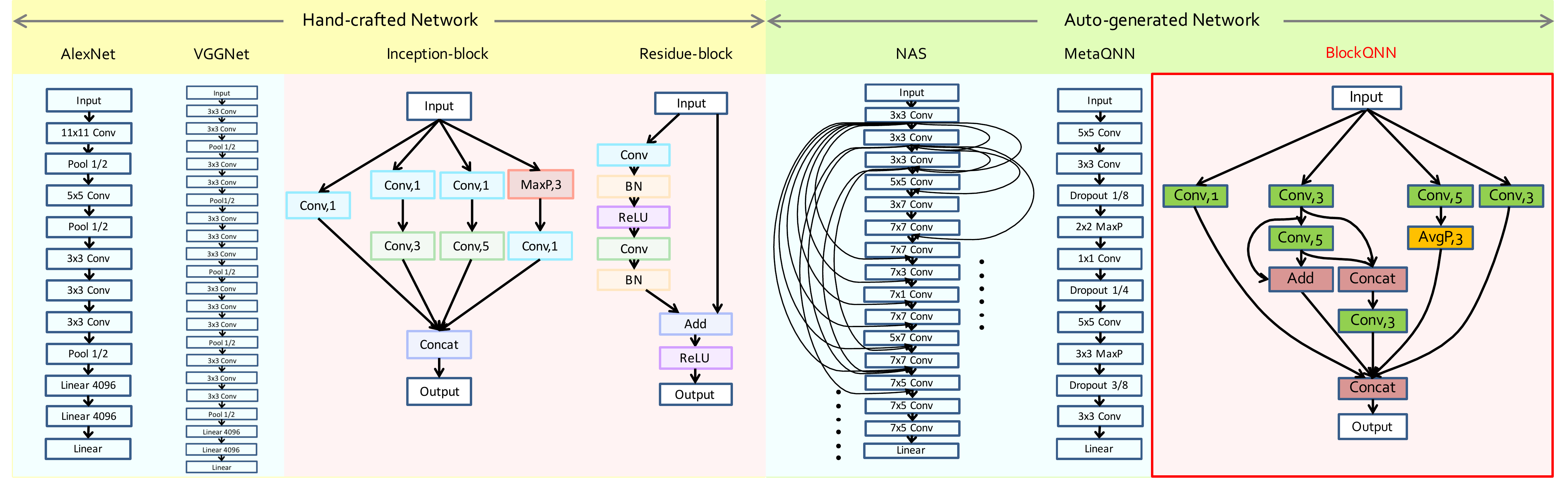}
	\caption{The proposed \textbf{BlockQNN} (right in red box) compared with the hand-crafted networks marked in yellow and the existing auto-generated networks in green. Automatically generating the \texttt{plain} networks~\cite{baker2016designing,zoph2016neural} marked in blue need large computational costs on searching optimal layer types and hyperparameters for each single layer, while the \texttt{block}-wise network heavily reduces the cost to search structures only for one block. The entire network is then constructed by stacking the generated blocks. Similar \texttt{block} concept has been demonstrated its superiority in hand-crafted networks, such as inception-block and residue-block marked in red.}
	\label{fig:motivation_compare}
\end{figure*}
%=========== fig: fig1 ==================
% The development history of convolutional neural network. From hand-crafted networks to automatic designed networks. Block refers to basic network sub-structures, such as inception module and residual module.}

% Modern neural networks can have hundreds of layers and each layer can have many options in layer type and hyper-parameters, which makes the network design space huge. To exploit the space more efficiently, we only design blocks in network architecture, instead of the whole network. Actually, most CNN architectures can be viewed as the stack of several basic sub-structures usually called 'block'. The blocks are staked repeatedly to build a deep network. For example, the popular CNN models like VGG~\cite{simonyan2014very}, Inception~\cite{szegedy2015going,szegedy2015rethinking} and Resnet~\cite{he2015deep} all have their own unique blocks. Due to the block-constructive architecture, these networks have powerful generalization ability that can transfer to different datasets and application areas.
%application domains???

% ========= Challenge of automatic model generation
Although some recent works have attempted computer-aided or automated network design~\cite{baker2016designing,zoph2016neural}, there are several challenges still unsolved: 
(1) Modern neural networks always consist of hundreds of convolutional layers, each of which has numerous options in type and hyperparameters.
It makes a huge search space and heavy computational costs for network generation.
(2) One typically designed network is usually limited on a specific dataset or task, and thus is hard to transfer to other tasks or generalize to another dataset with different input data sizes.
In this paper, we provide a solution to the aforementioned challenges by a novel fast Q-learning framework, called \textit{BlockQNN}, to automatically design the network architecture, as shown in Fig.~\ref{fig:motivation_compare}.

% ====== Method point1
Particularly, to make the network generation efficient and generalizable, we introduce the \texttt{block}-wise network generation,~\ie, we construct the network architecture as a flexible stack of personalized blocks rather tedious per-layer network piling.
Indeed, most modern CNN architectures such as Inception~\cite{szegedy2015going,ioffe2015batch,szegedy2015rethinking} and ResNet Series~\cite{he2015deep,he2016identity} are assembled as the stack of basic \texttt{block} structures.
For example, the inception and residual blocks shown in Fig.~\ref{fig:motivation_compare} are repeatedly concatenated to construct the entire network.
With such kind of \texttt{block}-wise network architecture, the generated network owns a powerful generalization to other task domains or different datasets.

% In this paper, we make efforts to design a CNN architecture in an automatic manner with a block-wise concept, efficiently and practically.

% ====== Method point2
% To overcome the greatest challenge of the huge demand of computing resource for this task, we focus on designing network blocks for building network, instead of generating whole network directly, to reduce the search space of network design.
% 
In comparison to previous methods like NAS~\cite{zoph2016neural} and MetaQNN~\cite{baker2016designing}, as depicted in Fig.~\ref{fig:motivation_compare}, we present a more readily and elegant model generation method that specifically designed for block-wise generation.
Motivated by the unsupervised reinforcement learning paradigm, we employ the well-known Q-learning~\cite{watkins1989learning} with experience replay~\cite{lin1993reinforcement} and epsilon-greedy strategy~\cite{mnih2015human} to effectively and efficiently search the optimal block structure.
The network block is constructed by the learning agent which is trained sequentiality to choose component layers. Afterwards we stack the block to construct the whole auto-generated network.
%
%and stacking configuration.
% 
%Moreover, to enable efficient search with fast convergence, we also propose an early stop strategy with a new reward function aiming at balancing the network performance and the network complexity.
Moreover, we propose an early stop strategy to enable efficient search with fast convergence. A novel reward function is designed to ensure the accuracy of the early stopped network having positive correlation with the converged network. We can pick up good blocks in reduced training time using this property.
With this acceleration strategy, we can construct a Q-learning agent to learn the optimal block-wise network structures for a given task with limited resources (\eg~few GPUs or short time period).
The generated architectures are thus succinct and have powerful generalization ability compared to the networks generated by the other automatic network generation methods.

The proposed \texttt{block}-wise network generation brings a few advantages as follows:
\begin{itemize}
	\vspace{-0.2cm}
	\item \textit{Effective}. The automatically generated networks present comparable performances to those of hand-crafted networks with human expertise.
	The proposed method is also superior to the existing works and achieves a state-of-the-art performance on CIFAR-$10$ with $3.54\%$ error rate.
	\vspace{-0.25cm}
	\item \textit{Efficient}. We are the first to consider \texttt{block}-wise setup in automatic network generation.
	Companied with the proposed early stop strategy, the proposed method results in a fast search process.
	The network generation for CIFAR task reaches convergence with only $32$ GPUs in $3$ days, which is much more efficient than that by NAS~\cite{zoph2016neural} with $800$ GPUs in $28$ days.
	\vspace{-0.25cm}
	\item \textit{Transferable}. It offers surprisingly superior transferable ability that the network generated for CIFAR can be transferred to ImageNet with little modification but still achieve outstanding performance.
\end{itemize}
%-------------------------------------------------------------------------

%%%%%%%%%%%%%%%%%%%%%%%%%%%%%%%%%%% RELATED WORK
\section{Related Work}
Early works, from $1980$s, have made efforts on automating neural network design which often searched good architecture by the genetic algorithm or other evolutionary algorithms~\cite{schaffer1992combinations,stanley2002evolving,stanley2009hypercube,suganuma2017genetic,saxena2016convolutional,domhan2015speeding,xie2017genetic}. Nevertheless, these works, to our best knowledge, cannot perform competitively compared with hand-crafted networks. 
% 
% Recent works, \ie~Neural Architecture Search (NAS)~\cite{zoph2016neural} and MetaQNN~\cite{baker2016designing}, based on reinforcement learning, provide powerful capability to beat the state-of-the-art hand-crafted networks.
% 
Recent works, \ie~Neural Architecture Search (NAS)~\cite{zoph2016neural} and MetaQNN~\cite{baker2016designing}, adopted reinforcement learning to automatically search a good network architecture. 
Although they can yield good performance on small datasets such as CIFAR-$10$, CIFAR-$100$, the direct use of MetaQNN or NAS for architecture design on big datasets like ImageNet~\cite{deng2009imagenet} is computationally expensive via searching in a huge space.
Besides, the network generated by this kind of methods is task-specific or dataset-specific, that is, it cannot been well transferred to other tasks nor datasets with different input data sizes. For example, the network designed for CIFAR-$10$ cannot been generalized to ImageNet.

%$t\in\mathbf{T}, (\mathbf{T}=\{1,2,3,...\})$
%$k,k^{'}\in \mathbf{K}, \mathbf{K}=\{1,2,...,t-1\}$.
%table 1
\begin{table}[t!]
	%% increase table row spacing, adjust to taste
	%\setlength{\tabcolsep}{3pt}
	\renewcommand{\arraystretch}{1.3}
	\label{table:code_space}
	\vspace{-0.1cm}
	%\centering
	\begin{center}
		\footnotesize
		\begin{tabular}{c|c|c|c|c|c}
			\hline
			Name&Index&Type&Kernel Size&Pred1&Pred2 \\
			\hline
			Convolution&\(\mathbf{T}\)&1& 1, 3, 5 & \(\mathbf{K}\) & 0 \\
			\hline
			Max Pooling&\(\mathbf{T}\)&2&1, 3 & \(\mathbf{K}\) & 0 \\
			\hline
			Average Pooling&\(\mathbf{T}\)&3& 1, 3 & \(\mathbf{K}\) & 0\\
			\hline
			Identity&\(\mathbf{T}\)&4& 0 & \(\mathbf{K}\) & 0\\
			\hline
			Elemental Add&\(\mathbf{T}\)&5& 0 & \(\mathbf{K}\) & \(\mathbf{K}\)\\
			\hline
			Concat&\(\mathbf{T}\)&6& 0 & \(\mathbf{K}\) & \(\mathbf{K}\)\\
			\hline
			Terminal&\(\mathbf{T}\)&7& 0 & 0 & 0\\
			\hline
		\end{tabular}
	\end{center}
	\caption{Network Structure Code Space. The space contains seven types of commonly used layers. Layer index stands for the position of the current layer in a block, the range of the parameters is set to be $\mathbf{T}=\{1,2,3,...\mbox{max layer index}\}$. Three kinds of kernel sizes are considered for convolution layer and two sizes for pooling layer. Pred$1$ and Pred$2$ refer to the predecessor parameters which is used to represent the index of layer’s predecessor, the allowed range is $\mathbf{K}=\{1,2,...,\mbox{current layer index}-1\}$}
	\vspace{-0.1cm}
\end{table}
Instead, our approach is aimed to design network block architecture by an efficient search method with a distributed asynchronous Q-learning framework as well as an early-stop strategy.
The block design conception follows the modern convolutional neural networks such as Inception~\cite{szegedy2015going,ioffe2015batch,szegedy2015rethinking} and Resnet~\cite{he2015deep,he2016identity}. The inception-based networks construct the \texttt{inception blocks} via a hand-crafted multi-level feature extractor strategy by computing $1\times 1$, $3\times 3$, and $5\times 5$ convolutions, while the Resnet uses \texttt{residue blocks} with shortcut connection to make it easier to represent the identity mapping which allows a very deep network. 
The \texttt{blocks} automatically generated by our approach have similar structures such as some blocks contain short cut connections and inception-like multi-branch combination. We will discuss the details in Section~\ref{subsec:block_analysis}.

Another bunch of related works include hyper-parameter optimization~\cite{bergstra2011algorithms}, meta-learning~\cite{vilalta2002perspective} and learning to learn methods~\cite{hochreiter2001learning,andrychowicz2016learning}. However, the goal of these works is to use meta-data to improve the performance of the existing algorithms, such as finding the optimal learning rate of optimization methods or the optimal number of hidden layers to construct the network. In this paper, we focus on learning the entire topological architecture of network blocks to improve the performance.

\begin{figure}[tbp]
	\centering
	\includegraphics[width=\linewidth]{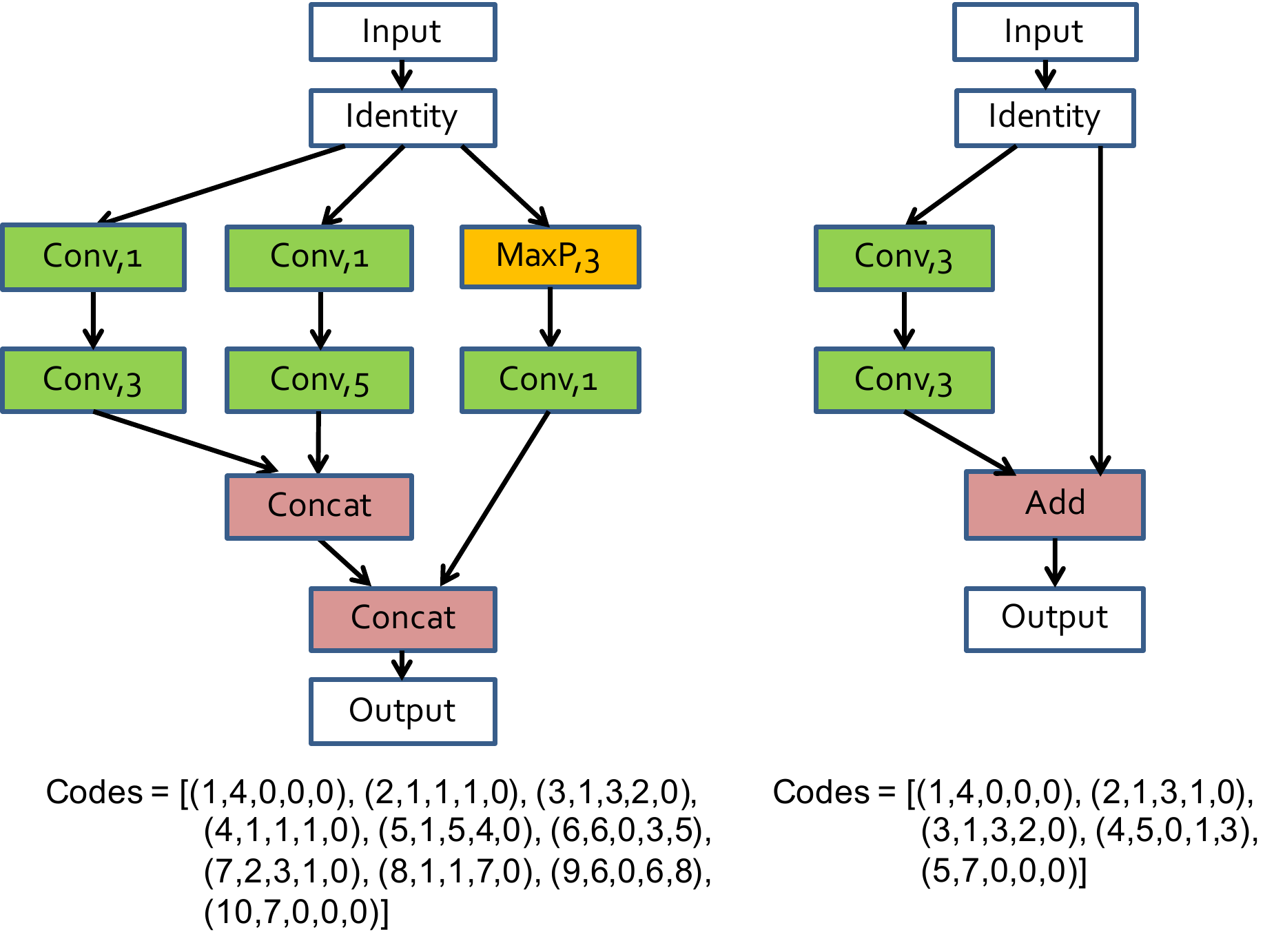}
	\caption{Representative block exemplars with their Network structure codes (NSC) respectively: the block with multi-branch connections (left) and the block with shortcut connections (right).}
	\vspace{-0.3cm}
	\label{fig:block_codes}
\end{figure}

%%%%%%%%%%%%%%%%%%%%%%%%%%%%%%%%%%% METHODOLOGY
\section{Methodology}
\label{sec:method}

\subsection{ Convolutional Neural Network Blocks}
\label{subsec:blocks}

The modern CNNs, \eg~Inception and Resnet, are designed by stacking several \texttt{blocks} each of which shares similar structure but with different weights and filter numbers to construct the network. With the block-wise design, the network can not only achieves high performance but also has powerful generalization ability to different datasets and tasks. Unlike previous research on automating neural network design which generate the entire network directly, we aim at designing the \texttt{block} structure.

\begin{figure}[tbp]
	%\linewidth=1
	\centering
	\includegraphics[width=\linewidth]{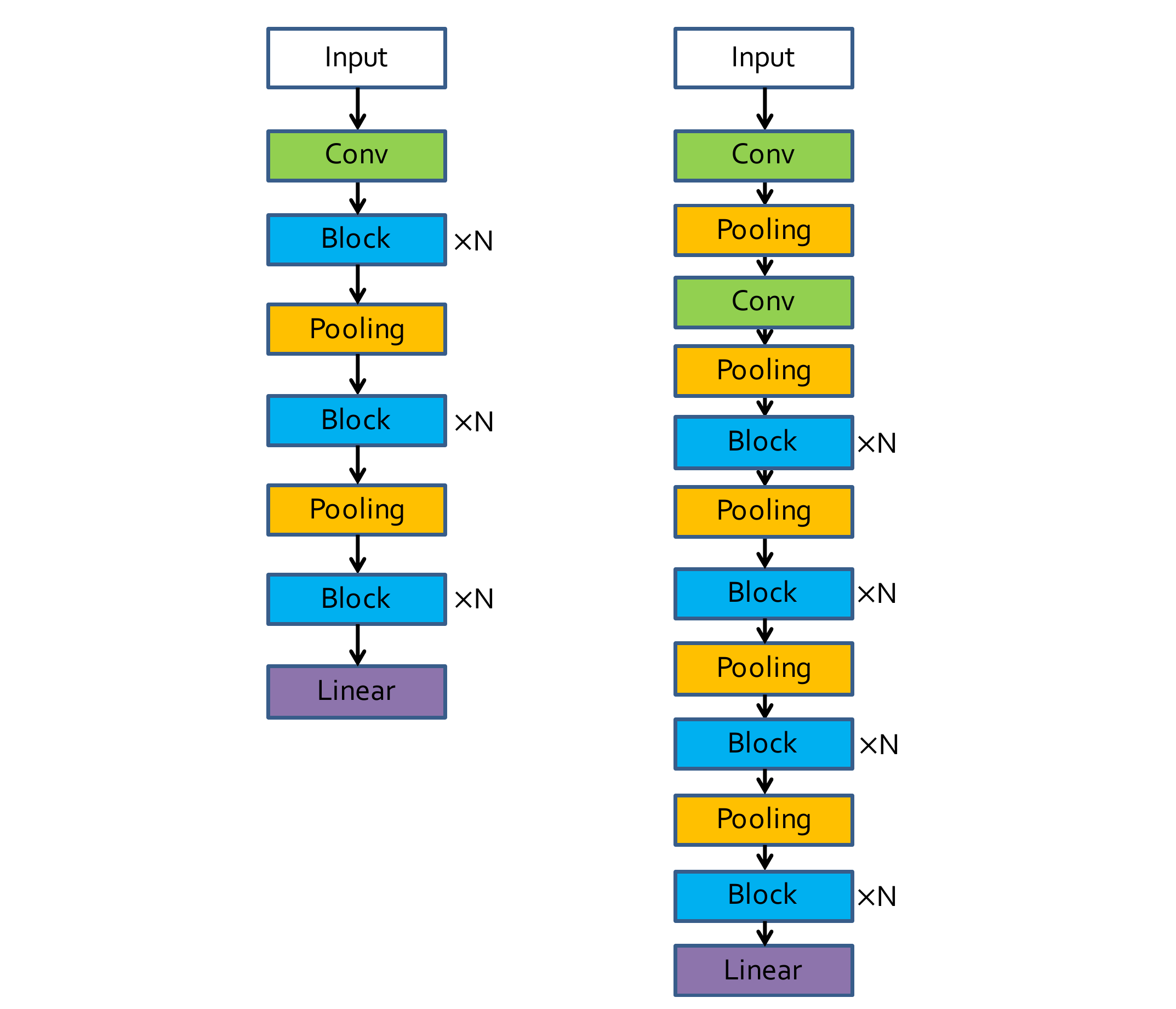}
	\caption{Auto-generated networks on CIFAR-$10$ (left) and ImageNet (right). Each network starts with a few convolution layers to learn low-level features, and followed by multiple repeated \texttt{blocks} with several pooling layers inserted to downsample.}
	\vspace{-0.3cm}
	\label{fig:architectures}
\end{figure}

\begin{figure*}[tbp]
	\centering
	\includegraphics[width=\linewidth]{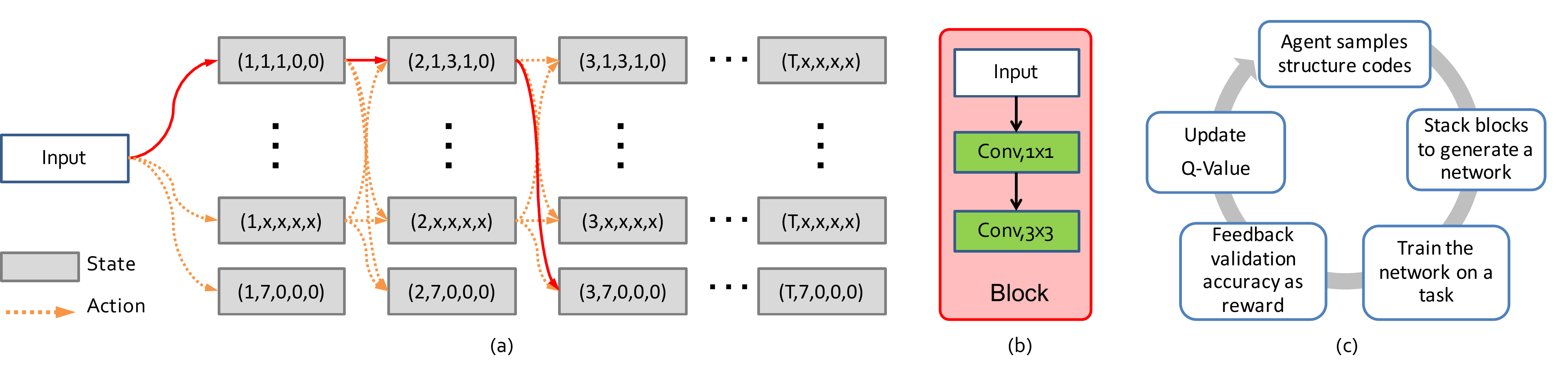}
	\caption{Q-learning process illustration. (a) The state transition process by different action choices. The block structure in (b) is generated by the red solid line in (a). (c) The flow chart of the Q-learning procedure.}
	\label{fig:state_action}
\end{figure*}

As a CNN contains a feed-forward computation procedure, we represent it by a directed acyclic graph (DAG), where each node corresponds to a layer in the CNN while directed edges stand for data flow from one layer to another. To turn such a graph into a uniform representation, we propose a novel layer representation called \textbf{Network Structure Code} (NSC), as shown in Table~\ref{table:code_space}. Each block is then depicted by a set of $5$-D NSC vectors. In NSC, the first three numbers stand for the \textit{layer index}, \textit{operation type} and \textit{kernel size}. The last two are \textit{predecessor parameters} which refer to the position of a layer's predecessor layer in structure codes. The second predecessor (Pred$2$) is set for the layer owns two predecessors, and for the layer with only one predecessor, Pred$2$ will be set to zero. This design is motivated by the current powerful hand-crafted networks like Inception and Resnet which own their special block structures. This kind of block structure shares similar properties such as containing more complex connections, \eg~shortcut connections or multi-branch connections, than the simple connections of the plain network like AlexNet. Thus, the proposed NSC can encode complexity architectures as shown in Fig.~\ref{fig:block_codes}. 
In addition, all of the layer without successor in the block are concatenated together to provide the final output.
Note that each convolution operation, same as the declaration in Resnet~\cite{he2016identity}, refers to a \textbf{Pre-activation Convolutional Cell} (PCC) with three components, \ie~\textit{ReLU}, \textit{Convolution} and \textit{Batch Normalization}.
This results in a smaller searching space than that with three components separate search, and hence with the PCC, we can get better initialization for searching and generating optimal block structure with a quick training process.
% 

% (a) Network architectures we used for classification. They only consist of normal convolution layers, pooling layers and repeated blocks. The top one is for ImageNet task and the bottom one for CIFAR-10. 

Based on the above defined \texttt{blocks}, we construct the complete network by stacking these block structures sequentially which turn a common plain network into its counterpart block version.
Two representative auto-generated networks on CIFAR and ImageNet tasks are shown in Fig.~\ref{fig:architectures}. 
There is no down-sampling operation within each block. We perform down-sampling directly by the pooling layer. If the size of feature map is halved by pooling operation, the block's weights will be doubled.
% All blocks have the same feature map size between the pooling layers, and after pooling layer with a down sampling operation stride of two, feature map will be reduced by two but the block's weights will be doubled. here is no down sampling operation in block.
% 
The architecture for ImageNet contains more pooling layers than that for CIFAR because of their different input sizes, \ie~$224\times 224$ for ImageNet and $32\times 32$ for CIFAR.
More importantly, the \texttt{blocks} can be repeated any $N$ times to fulfill different demands, and even place the blocks in other manner, such as inserting the block into the Network-in-Network~\cite{lin2013network} framework or setting short cut connection between different blocks.

\subsection{Designing Network Blocks With Q-Learning}
\label{subsec:q_learning}

Albeit we squeeze the search space of the entire network design by focusing on constructing network \texttt{blocks}, there is still a large amount of possible structures to seek. Therefore, we employ reinforcement learning rather than random sampling for automatic design. Our method is based on Q-learning, a kind of reinforcement learning, which concerns how an agent ought to take actions so as to maximize the cumulative reward. The Q-learning model consists of an \textit{agent}, \textit{states} and a set of \textit{actions}. 

In this paper, the \textit{state} $s\in S$ represents the status of the current layer which is defined as a Network Structure Code (NSC) claimed in Section~\ref{subsec:blocks}, \ie~$5$-D vector \{layer index, layer type, kernel size, pred$1$, pred$2$\}. The \textit{action} $a\in A$ is the decision for the next successive layer. 
Thanks to the defined NSC set with a limited number of choices, both the \textit{state} and \textit{action} space are thus finite and discrete to ensure a relatively small searching space. 
The state transition process $(s_t,a(s_t))\rightarrow (s_{t+1})$ is shown in Fig.~\ref{fig:state_action}(a), where $t$ refers to the current layer. The block example in Fig.~\ref{fig:state_action}(b) is generated by the red solid lines in Fig.~\ref{fig:state_action}(a).
% 
% In addition, we restrict the environment to have a discrete and finite state space \(S\) as well as action space \(A\) to ensure a relatively small searching space. Figure. \ref{fig:state_action}(a) shows the state \(s \in S\), the action \(a \in A\)  and state transition process\((s_t,A(s_t)) \rightarrow (s_{t+1})\). 
% 
The learning agent is given the task of sequentially picking NSC of a block. The structure of block can be considered as an action selection trajectory $\tau_{a_{1:T}}$, \ie~a sequence of NSCs. 
We model the layer selection process as a Markov Decision Process with the assumption that a well-performing layer in one block should also perform well in another block~\cite{baker2016designing}.
To find the optimal architecture, we ask our agent to maximize its expected \textit{reward} over all possible trajectories, denoted by \(R_{\tau}\),
\begin{eqnarray}
R_{\tau}= \mathbb{E}_{P(\tau_{a_{1:T}})}[\mathbb{R}],
\end{eqnarray}
where the \(\mathbb{R}\) is the cumulative reward. For this maximization problem, it is usually to use recursive Bellman Equation to optimality. Given a state \(s_t\in S\), and subsequent action \(a\in A(s_t)\), we define the maximum total expected reward to be \(Q^*(s_t,a)\) which is known as Q-value of state-action pair. The recursive Bellman Equation then can be written as 
\begin{eqnarray}
\nonumber Q^*(s_t,a)=\mathbb{E}_{s_{t+1}|s_t,a}[\mathbb{E}_{r|s_t,a,s_{t+1}}[r|s_t,a,s_{t+1}]\\
+\gamma \max_{a'\in A(s_{t+1}))}Q^*(s_{t+1},a')].
\end{eqnarray}

%An empirical iterative approximation of the above quantity is
%it can be formulated as an iterative update
An empirical assumption to solve the above quantity is to formulate it as an iterative update:
\begin{align}
Q(s_T,a) =& 0\\
Q(s_{T-1},a_T) =& (1-\alpha)Q(s_{T-1},a_T) + \alpha r_T\\
\nonumber Q(s_t,a)=&(1-\alpha)Q(s_t,a)\\
+ \alpha [r_t+\gamma \max_{a'}&Q(s_{t+1},a')], t \in\{1,2,...T-2\},
\end{align}
%\begin{multline}
%Q_{i+1}(s_t,a)=Q_{i}(s_t,a) +\\
%\alpha [r_t+\gamma \max_{a'\in A(s_{t+1}))}Q_i(s_{t+1},a')-Q_{i}(s_t,a)],
%\end{multline}
%for example, we can design a function to describe every layer's contribution for the performance of block.
where \(\alpha\) is the learning rate which determines how the newly acquired information overrides the old information, \(\gamma\) is the discount factor which measures the importance of future rewards.
%$r_T$ is the validation accuracy of corresponding network trained convergence on training set for final state $s_{T}$, \ie~terminal layers.
\(r_t\) denotes the intermediate reward observed for the current state \(s_{t}\) and $s_{T}$ refers to final state,~\ie terminal layers. $r_T$ is the validation accuracy of corresponding network trained convergence on training set for $a_T$,~\ie action to final state. Since the reward \(r_t\) cannot be explicitly measured in our task, we use reward shaping~\cite{ng1999policy} to speed up training. The shaped intermediate reward is defined as:
%we use the validation accuracy with the hyperparameter $\lambda$ as an alternative:
% and the Q-value is reformulated as
%\begin{multline}
%Q_{i+1}(s_t,a) = Q_{i}(s_t,a) +\\
%\alpha[\text{accuracy} +\gamma \max_{{a}'\in A(s_{t+1})}Q_i(s_{t+1},{a}')-Q_{i}(s_t,a)].
%\end{multline}
\begin{eqnarray}
r_t = \frac{r_T}{T}.
\end{eqnarray}

\begin{figure}[tbp]
	\centering
	\includegraphics[width=\linewidth]{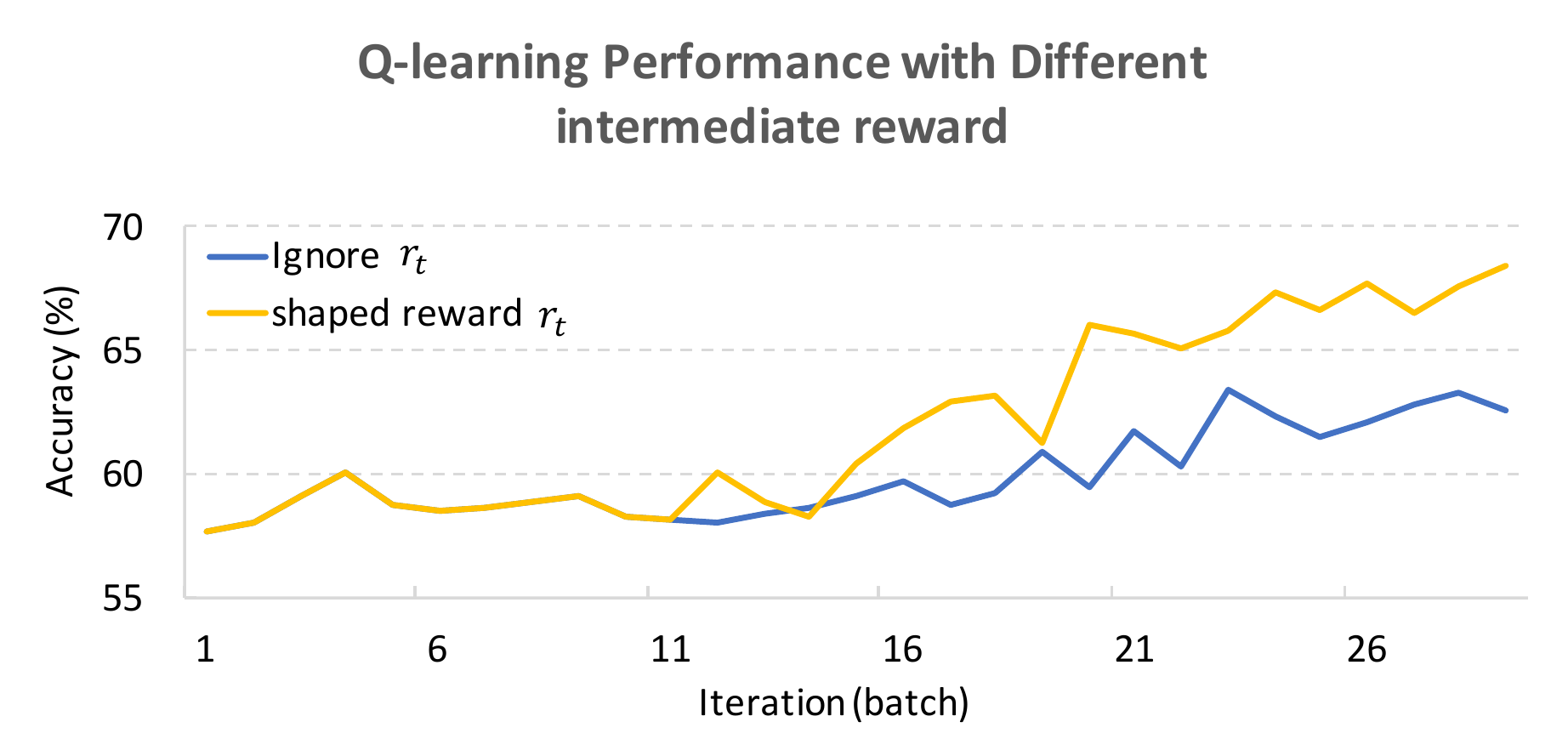}
	\caption{Comparison results of Q-learning with and without the shaped intermediate reward $r_t$. By taking our shaped reward, the learning process convergent faster than that without shaped reward start from the same exploration.}
	\label{fig:acc_reward}
\end{figure}

Previous works~\cite{baker2016designing} ignore these rewards in the iterative process, \ie~ set them to zero, which may cause a slow convergence in the beginning. This is known as the temporal credit assignment problem which makes RL time consuming~\cite{sutton1998reinforcement}. In this case, the Q-value of $s_T$ is much higher than others in early stage of training and thus leads the agent prefer to stop searching at the very beginning, \ie~tend to build small block with fewer layers.
%\begin{eqnarray}
%Q(s_t,a)\sim \alpha^{f-t}R\ll Q(s_T,a),  \alpha=0.01
%\end{eqnarray}
We show a comparison result in Fig.~\ref{fig:acc_reward}, the learning process of the agent with our shaped reward \(r_t\) convergent much faster than previous method.

% Q-learning result with different version of  Bellman¡¯s Equation approximation. The learning process with the \(r_t\), we use test accuracy to represent, can convergence faster than ignore it.

We summarize the learning procedure in Fig.~\ref{fig:state_action}(c). The agent first samples a set of structure codes to build the block architecture, based on which the entire network is constructed by stacking these blocks sequentially. We then train the generated network on a certain task, and the validation accuracy is regarded as the reward to update the Q-value. Afterwards, the agent picks another set of structure codes to get a better block structure.

\subsection{Early Stop Strategy}
\label{subsec:early_stop}

\begin{figure}[tbp]
	\centering
	\includegraphics[width=\linewidth]{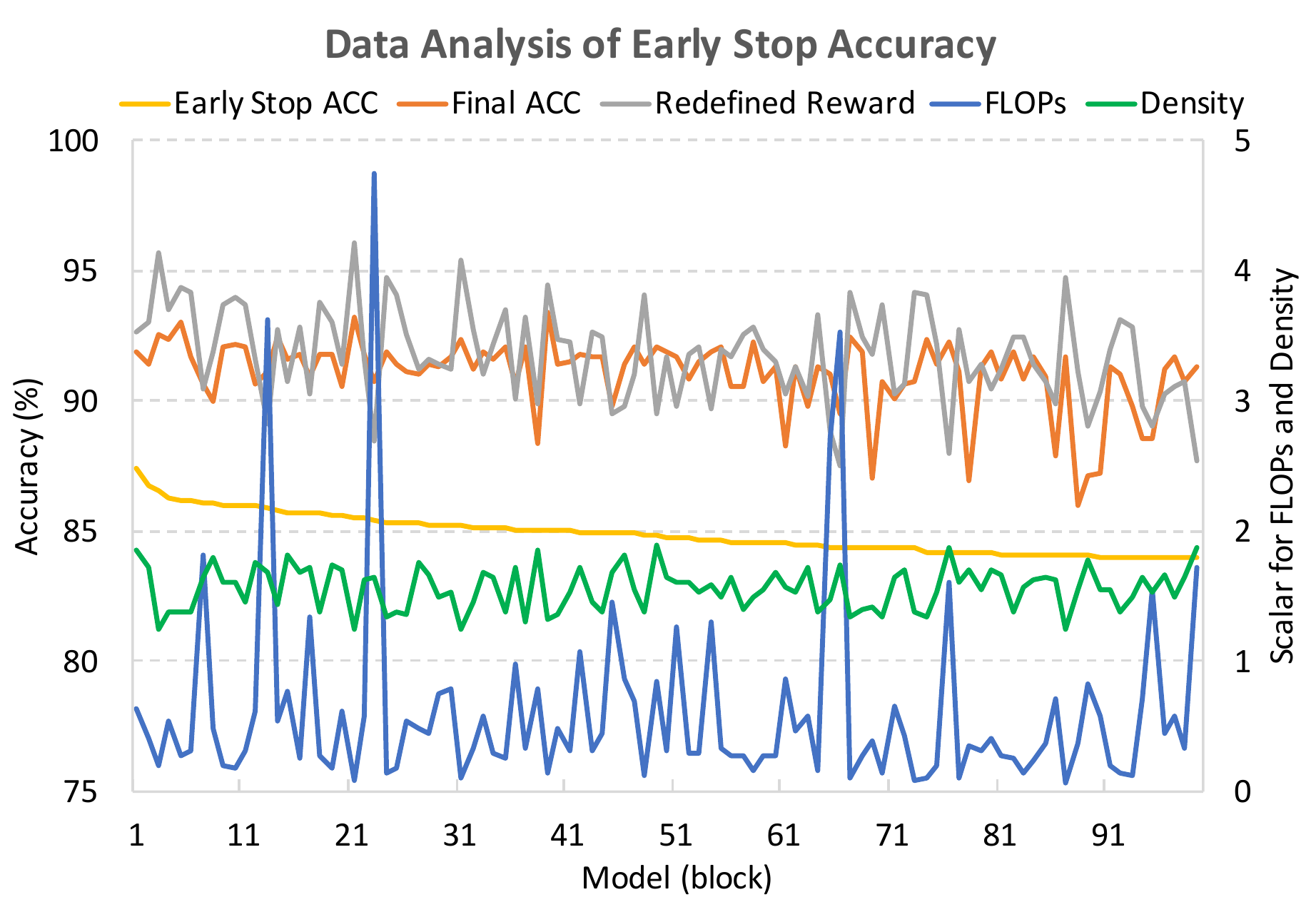}
	\caption{The performance of early stop training is poorer than the final accuracy of a complete training. With the help of FLOPs and Density, it squeezes the gap between the redefined reward function and the final accuracy.}
	\label{fig:early_stop}
\end{figure}

Introducing \texttt{block}-wise generation indeed increases the efficiency. However, it is still time consuming to complete the search process. To further accelerate the learning process, we introduce an early stop strategy. As we all know, early stopping training process might result in a poor accuracy. Fig.~\ref{fig:early_stop} shows an example, where the early-stop accuracy in yellow line is much lower than the final accuracy in orange line, which means that some good blocks unfortunately perform worse than bad blocks when stop training early. In the meanwhile, we notice that the FLOPs and density of the corresponding blocks have a negative correlation with the final accuracy. Thus, we redefine the reward function as 
\begin{eqnarray}
\nonumber reward = \text{ACC}_{\text{EarlyStop}} - \mu\log(\text{FLOPs})\\
- \rho\log(\text{Density}),
\label{eq:early_stop_reward}
\end{eqnarray}
where FLOPs~\cite{he2015convolutional} refer to an estimation of computational complexity of the block, and Density is the edge number divided by the dot number in DAG of the block. There are two hyperparameters, $\mu$ and $\rho$, to balance the weights of FLOPs and Density. With the redefined reward function, the reward is more relevant to the final accuracy.

With this early stop strategy and small searching space of network blocks, it just costs $3$ days to complete the searching process with only $32$ GPUs, which is superior to that of~\cite{zoph2016neural}, spends $28$ days with $800$ GPUs to achieve the same performance.

\section{Framework and Training Details}

\subsection{Distributed Asynchronous Framework}

\begin{figure}[tbp]
	\centering
	\includegraphics[width=\linewidth]{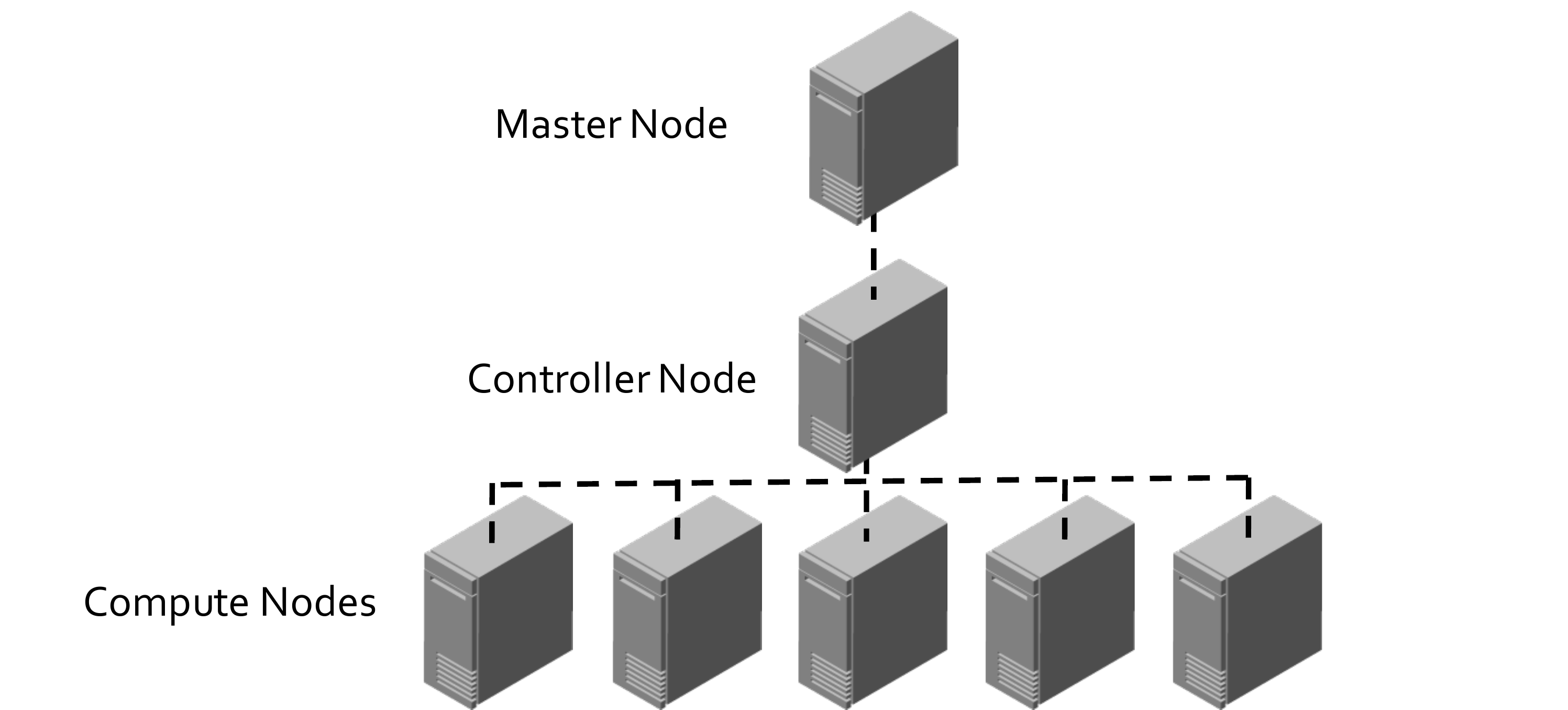}
	\caption{The distributed asynchronous  framework. It contains three parts: master node, controller node and compute nodes.}\label{fig:ps}
\end{figure}

To speed up the learning of agent, we use a distributed asynchronous framework as illustrated in Fig.~\ref{fig:ps}. It consists of three parts: master node, controller node and compute nodes. The agent first samples a batch of block structures in master node. Afterwards, we store them in a controller node which uses the block structures to build the entire networks and allocates these networks to compute nodes. 
It can be regarded as a simplified parameter-server~\cite{dean2012large,li2013parameter}. Specifically, the network is trained in parallel on each of compute nodes and returns the validation accuracy as reward by controller nodes to update agent. With this framework, we can generate network efficiently on multiple machines with multiple GPUs.

\subsection{Training Details}

\vspace{0.1cm}\noindent \textbf{Epsilon-greedy Strategy.} The agent is trained using Q-learning with experience replay~\cite{lin1993reinforcement} and epsilon-greedy strategy~\cite{mnih2015human}. %Table~\ref{table:2} shows the epsilon schedule we used in the experiments. 
With epsilon-greedy strategy, the random action is taken with probability $\epsilon$ and the greedy action is chosen with probability $1-\epsilon$. We decrease epsilon from $1.0$ to $0.1$ following the epsilon schedule as shown in Table~\ref{table:2} such that the agent can transform smoothly from exploration to exploitation. We find that the result goes better with a longer exploration, since the searching scope would become larger and the agent can see more block structures in the random exploration period. 
%
%Table~\ref{table:2} shows that the number of batches for different epsilon we used in the experiments. We decrease epsilon from 1.0 to 0.1, leaving half the searching time for exploration.
%

\vspace{0.1cm}\noindent\textbf{Experience Replay.} Following~\cite{baker2016designing}, we employ a replay memory to store the validation accuracy and block description after each iteration. Within a given interval, \ie~each training iteration, the agent samples $64$ blocks with their corresponding validation accuracies from the memory and updates Q-value $64$ times.
%table 2
\begin{table}[t!]
	%% increase table row spacing, adjust to taste
	\renewcommand{\arraystretch}{1.3}
	%\centering
	\begin{center}
		\scriptsize
		\begin{tabular}{c|c|c|c|c|c|c|c|c|c|c}
			\hline
			$\epsilon$&1.0& 0.9 & 0.8 & 0.7 & 0.6 & 0.5 & 0.4 & 0.3 & 0.2 & 0.1 \\
			\hline
			Iters&95& 7 & 7 & 7 & 10 & 10 & 10 & 10 & 10 & 12 \\
			\hline
		\end{tabular}
	\end{center}
	\caption{Epsilon Schedules. The number of iteration the agent trains at each epsilon($\epsilon$) state.}\label{table:2}
\end{table}

\vspace{0.1cm}\noindent \textbf{BlockQNN Generation.} 
In the Q-learning update process, the learning rate \(\alpha\) is set to 0.01 and the discount factor \(\gamma\) is $1$. We set the hyperparameters $\mu$ and $\rho$ in the redefined reward function as $1$ and $8$, respectively. The agent samples $64$ sets of NSC vectors at a time to compose a mini-batch and the maximum layer index for a block is set to $23$. We train the agent with $178$ iterations, \ie~sampling $11,392$ blocks in total. 

During the block searching phase, the compute nodes train each generated network for a fixed $12$ epochs on CIFAR-$100$ using the early top strategy as described in Section~\ref{subsec:early_stop}. CIFAR-100 contains $60,000$ samples with $100$ classes which are divided into training and test set with the ratio of $5:1$. We train the network without any data augmentation procedure. 
The batch size is set to 256. We use Adam optimizer~\cite{kingma2014adam} with \(\beta_1 = 0.9  \), \(\beta_2 = 0.999  \), \(\varepsilon =10^{-8} \). The initial learning rate is set to 0.001 and is reduced with a factor of $0.2$ every $5$ epochs. All weights are initialized as in~\cite{he2015delving}. 
If the training result after the first epoch is worse than the random guess, we reduce the learning rate by a factor of $0.4$ and restart training, with a maximum of $3$ times for restart-operations. 

After obtaining one optimal block structure, we build the whole network with stacked blocks and train the network until converging to get the validation accuracy as the criterion to pick the best network. In this phase, we augment data with randomly cropping the images with size of $32\times 32$ and horizontal flipping. All models use the SGD optimizer with momentum rate set to $0.9$ and weight decay set to $0.0005$. We start with a learning rate of $0.1$ and train the models for $300$ epochs, reducing the learning rate in the $150$-th and $225$-th epoch. The batch size is set to $128$ and all weights are initialized with MSRA initialization~\cite{he2015delving}.

\vspace{0.1cm}\noindent \textbf{Transferable BlockQNN.} 
We also evaluate the transferability of the best auto-generated block structure searched on CIFAR-$100$ to a smaller dataset, CIFAR-$10$, with only $10$ classes and a larger dataset, ImageNet, containing $1.2$M images with $1000$ classes. All the experimental settings are the same as those on the CIFAR-$100$ stated above.
The training is conducted with a mini-batch size of $256$ where each image has data augmentation of randomly cropping and flipping, and is optimized with SGD strategy. The initial learning rate, weight decay and momentum are set as $0.1$, $0.0001$ and $0.9$, respectively. We divide the learning rate by $10$ twice, at the $30$-th and $60$-th epochs. The network is trained with a total of $90$ epochs. We evaluate the accuracy on the test images with center crop.

%\color{cyan} [not necessary?? if space is not enough, suggest remove this paragraph.] 
Our framework is implemented under the PyTorch scientific computing platform. We use the CUDA backend and cuDNN accelerated library in our implementation for high-performance GPU acceleration. Our experiments are carried out on $32$ NVIDIA TitanX GPUs and took about $3$ days to complete searching.

\begin{figure*}[tbp]
	\centering
	\includegraphics[width=\linewidth]{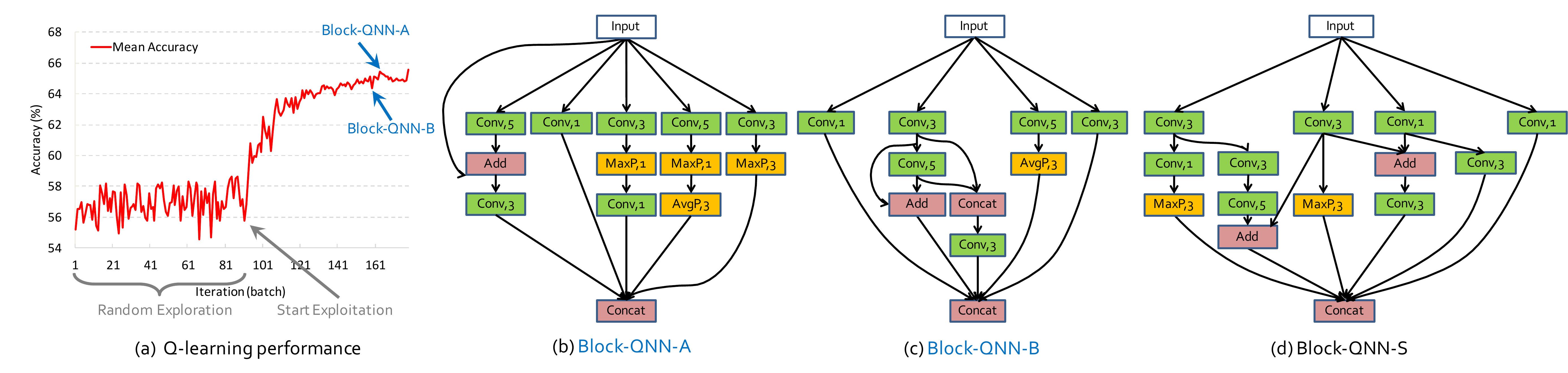}
	\vspace{0.05cm}
	\caption{(a) Q-learning performance on CIFAR-$100$. The accuracy goes up with the epsilon decrease and the top models are all found in the final stage, show that our agent can learn to generate better block structures instead of random searching. (b-c) Topology of the Top-$2$ block structures generated by our approach. We call them Block-QNN-A and Block-QNN-B. (d) Topology of the best block structures generated with limited parameters, named Block-QNN-S.  }
	\label{fig:8}
\end{figure*}

\begin{figure}[tbp]
	\centering
	\includegraphics[width=\linewidth]{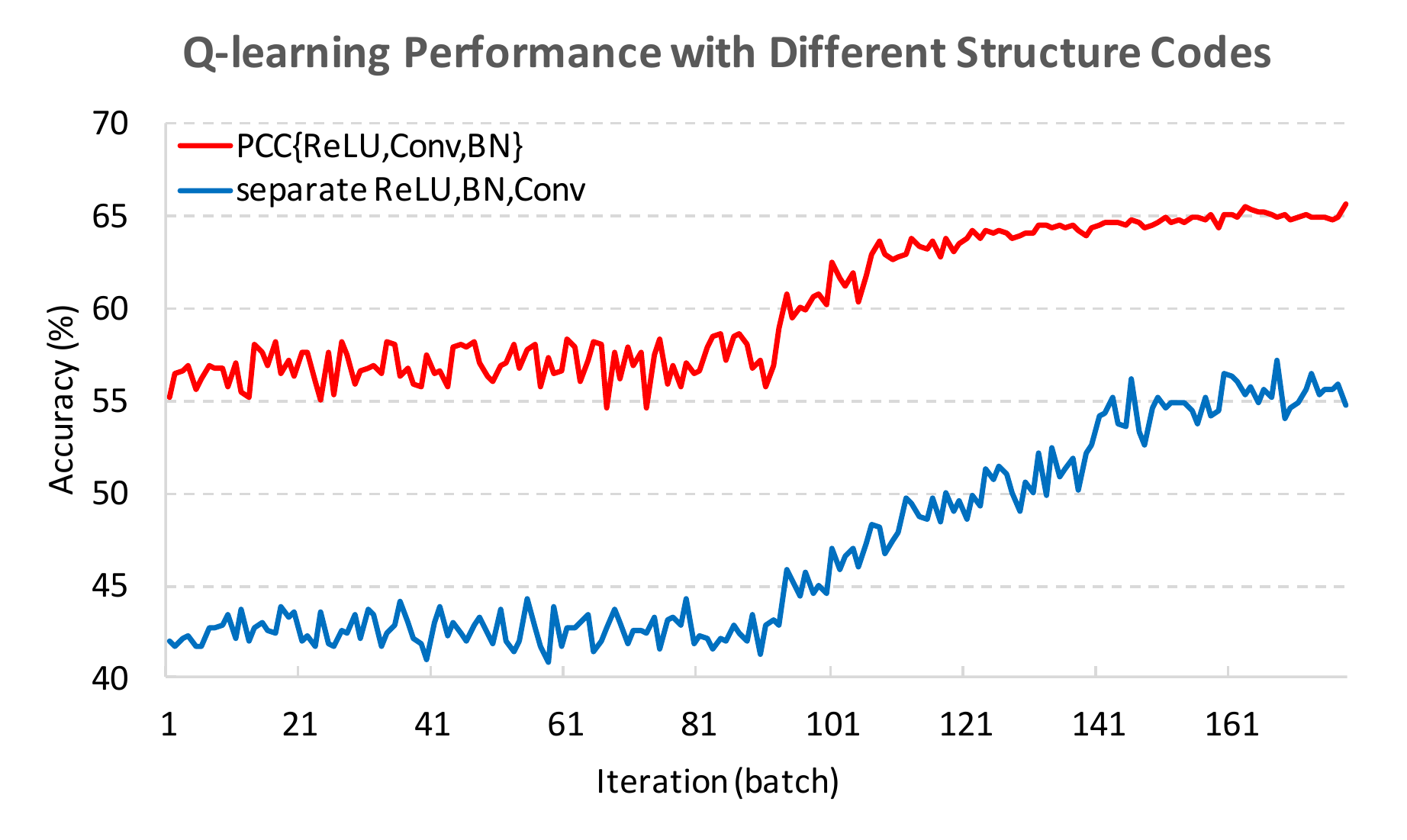}
	\caption{Q-learning result with different NSC on CIFAR-$100$. The red line refers to searching with PCC, \ie~combination of ReLU, Conv and BN. The blue stands for separate searching with ReLU, BN and Conv. The red line is better than blue from the beginning with a big gap.}\label{fig:pcc}
\end{figure}

\section{Results}

\subsection{Block Searching Analysis}
\label{subsec:block_analysis}
Fig.~\ref{fig:8}(a) provides early stop accuracies over $178$ batches on CIFAR-$100$, each of which is averaged over $64$ auto-generated block-wise network candidates within in each mini-batch. After random exploration, the early stop accuracy grows steadily till converges. The mean accuracy within the period of random exploration is $56\%$ while finally achieves $65\%$ in the last stage with $\epsilon=0.1$.
We choose top-$100$ block candidates and train their respective networks to verify the best block structure. 
We show top-$2$ block structures in Fig.~\ref{fig:8}(b-c), denoted as \textbf{Block-QNN-A} and \textbf{Block-QNN-B}.
%$\mathcal{B}_a$ and $\mathcal{B}_b$.
% 
% With the block structure, we can construct the network for different datasets and tasks easily.
% 
As shown in Fig.~\ref{fig:8}(a), both top-$2$ blocks are found in the final stage of the Q-learning process, which proves the effectiveness of the proposed method in searching optimal block structures rather than randomly searching a large amount of models.
Furthermore, we observe that the generated blocks share similar properties with those state-of-the-art hand-crafted networks. For example, Block-QNN-A and Block-QNN-B contain short-cut connections and multi-branch structures which have been manually designed in residual-based and inception-based networks. Compared to other auto-generated methods, the networks generated by our approach are more elegant and can automatically and effectively reveal the beneficial properties for optimal network structure.

To squeeze the searching space, as stated in Section~\ref{subsec:blocks}, we define a Pre-activation Convolutional Cell (PCC) consists of three components, \ie~ReLU, convolution and batch normalization (BN). We show the superiority of the PCC, searching a combination of three components, in Fig.~\ref{fig:pcc}, compared to the separate search of each component. Searching the three components separately is more likely to generate ``bad'' blocks and also needs more searching space and time to pursue ``good'' blocks.

%table 3
\begin{table}[t!]
	%% increase table row spacing, adjust to taste
	\renewcommand{\arraystretch}{1.3}
	%\centering
	
	\begin{center}
		\footnotesize
		\begin{tabular}{c|c|c|c|c}
			\hline
			Method  & Depth &Para &C-10  &C-100 \\
			\hline
			\hline
			%Network in Network~\cite{lin2013network} & - & 8.81 & 35.68 \\
			%\hline
			%Highway Network~\cite{srivastava2015highway} & - & 7.72 & -  \\
			%\hline
			%All-CNN~\cite{springenberg2014striving} & - & 7.25 & 33.71  \\
			%\hline
			VGG~\cite{simonyan2014very} & - & &7.25 & -  \\
			\hline
			\hline
			ResNet~\cite{he2015deep} & 110 &1.7M &6.61 & -  \\
			\hline
			Wide ResNet~\cite{zagoruyko2016wide} & 28 & 36.5M&4.17 & 20.5  \\
			\hline
			ResNet (pre-activation)~\cite{he2016identity} & 1001 &10.2M &4.62 & 22.71 \\
			\hline
			DenseNet (k = 12) ~\cite{huang2016densely} & 40 & 1.0M&5.24 & 24.42 \\
			DenseNet (k = 12)~\cite{huang2016densely} & 100 &7.0M &4.10 & 20.20 \\
			DenseNet (k = 24)~\cite{huang2016densely} & 100 &27.2M &3.74 & 19.25 \\
			DenseNet-BC (k = 40)~\cite{huang2016densely} & 190 &25.6M &3.46 & 17.18  \\
			\hline
			\hline
			MetaQNN (ensemble)~\cite{baker2016designing} & - & - &7.32 & - \\
			MetaQNN (top model)~\cite{baker2016designing} & - & 11.2M &6.92 & 27.14 \\
			\hline
			NAS v1~\cite{zoph2016neural} & 15 & 4.2M&5.50 & - \\
			NAS v2~\cite{zoph2016neural} & 20 &2.5M &6.01 & - \\
			NAS v3~\cite{zoph2016neural} & 39 &7.1M &4.47 & - \\
			NAS v3 more filters~\cite{zoph2016neural} & 39 &37.4M &3.65 & - \\
			\hline
			Block-QNN-A, N=4 & 25 &-&3.60 &18.64  \\
			Block-QNN-B, N=4 & 37 &-&3.80 & 18.72\\
			Block-QNN-S, N=2 & 19 &6.1M & 4.38 & 20.65 \\
			Block-QNN-S more filters &22&39.8M&3.54&18.06\\
			\hline
		\end{tabular}
	\end{center}
	\caption{Block-QNN's results (error rate) compare with state-of-the-art methods on CIFAR-$10$ (C-$10$) and CIFAR-$100$ (C-$100$) dataset. }\label{table:3}
\end{table}

\subsection{Results on CIFAR}

Due to the small size of images (\ie~$32\times 32$) in CIFAR, we set block stack number as $N=4$. We compare our generated best architectures with the state-of-the-art hand-crafted networks or auto-generated networks in Table~\ref{table:3}.

\vspace{0.1cm} \noindent \textit{Comparison with hand-crafted networks -}
It shows that our Block-QNN networks outperform most hand-crafted networks. The DenseNet-BC~\cite{huang2016densely} uses additional $1\times 1$ convolutions in each composite function and compressive transition layer to reduce parameters and improve performance, which is not adopted in our design. Our performance can be further improved by using this prior knowledge.

%that we do not introduce extra prior knowledge during training.

\vspace{0.1cm} \noindent \textit{Comparison with auto-generated networks -}
Our approach achieves a significant improvement to the MetaQNN~\cite{baker2016designing}, and even better than NAS's best model (\ie~\textit{NASv3 more filters})~\cite{zoph2016neural} proposed by Google brain which needs an expensive costs on time and GPU resources. As shown in Table~\ref{table:4}, NAS trains the whole system on $\mathbf{800}$ \textbf{GPUs} in $\mathbf{28}$ \textbf{days} while we only need $\mathbf{32}$ \textbf{GPUs} in $\mathbf{3}$ \textbf{days} to get state-of-the-art performance.

\vspace{0.1cm} \noindent \textit{Transfer block from CIFAR-$100$ to CIFAR-$10$ -}
We transfer the top blocks learned from CIFAR-$100$ to CIFAR-$10$ dataset, all experiment settings are the same. As shown in Table~\ref{table:3}, the blocks can also achieve state-of-the-art results on CIFAR-$10$ dataset with $3.60\%$ error rate that proved Block-QNN networks have powerful transferable ability.

\vspace{0.1cm} \noindent \textit{Analysis on network parameters -}
The networks generated by our method might be complex with a large amount of parameters since we do not add any constraints during training. We further conduct an experiment on searching networks with limited parameters and adaptive block numbers. We set the maximal parameter number as $10$M and obtain an optimal block (\ie~\textbf{Block-QNN-S}) which outperforms NASv3 with less parameters, as shown in Fig.~\ref{fig:8}(d). In addition, when involving more filters in each convolutional layer (\eg~from [$32$,$64$,$128$] to [$80$,$160$,$320$]), we can achieve even better result ($3.54\%$).

\subsection{Transfer to ImageNet}
To demonstrate the generalizability of our approach, we transfer the block structure learned from CIFAR to ImageNet dataset. 

For the ImageNet task, we set block repeat number $N=3$ and add more down sampling operation before blocks, the filters for convolution layers in different level blocks are [$64$,$128$,$256$,$512$]. We use the best blocks structure learned from CIFAR-$100$ directly without any fine-tuning, and the generated network initialized with MSRA initialization as same as above. The experimental results are shown in Table~\ref{table:5}. The network generated by our framework can get competitive result compared with other human designed models. The recently proposed methods such as Xception~\cite{chollet2016xception} and ResNext~\cite{xie2016aggregated} use special depth-wise convolution operation to reduce their total number of parameters and to improve performance. In our work, we do not use this new convolution operation, so it can't be compared fairly, and we will consider this in our future work to further improve the performance.

%table 4
\begin{table}[t!]
	%% increase table row spacing, adjust to taste
	\renewcommand{\arraystretch}{1.3}
	%\centering
	\footnotesize
	\begin{center}
		\begin{tabular}{c|c|c|c}
			\hline
			Method  & Best Model on CIFAR10  &GPUs & Time(days)\\
			\hline
			
			\hline
			MetaQNN~\cite{baker2016designing} & 6.92 & 10 & 10 \\
			\hline
			NAS~\cite{zoph2016neural} & 3.65 & 800 & 28  \\
			\hline
			
			\hline
			Our approach & 3.54 & 32 & 3  \\
			\hline
		\end{tabular}
	\end{center}
	\caption{The required computing resource and time of our approach compare with other automatic designing network methods.}\label{table:4}
\end{table}

%table 4
\begin{table}[t!]
	%% increase table row spacing, adjust to taste
	\renewcommand{\arraystretch}{1.3}
	%\centering
	\footnotesize
	\begin{center}
		\begin{tabular}{c|c|c|c|c}
			\hline
			Method  & Input Size &Depth & Top-1& Top-5\\
			\hline
			
			\hline
			VGG~\cite{simonyan2014very} & 224x224 & 16 & 28.5 & 9.90  \\
			\hline
			Inception V1~\cite{szegedy2015going} & 224x224 & 22& 27.8 & 10.10  \\
			%\hline
			Inception V2~\cite{ioffe2015batch} & 224x224 & 22& 25.2 & 7.80  \\
			\hline
			ResNet-50~\cite{he2016identity}  & 224x224& 50 &  24.7 & 7.80 \\
			%\hline
			ResNet-152~\cite{he2016identity} & 224x224&  152 &23.0 & 6.70 \\
			\hline
			Xception(our test)~\cite{chollet2016xception}  & 224x224& 50 &  23.6 & 7.10 \\
			%\hline
			ResNext-101(64x4d)~\cite{xie2016aggregated} & 224x224&  101 & 20.4 & 5.30 \\
			\hline
			
			\hline
			%Block-QNN-A  & 224x224 &  &  \\
			Block-QNN-B, N=3 & 224x224 & 38  &24.3  & 7.40  \\
			%Block-QNN-S, N=3 & 224x224 & 38  &22.9  & 6.56  \\
			Block-QNN-S, N=3 & 224x224 & 38  &22.6  & 6.46  \\
			\hline
		\end{tabular}
	\end{center}
	%\vspace{0.2cm}
	\caption{Block-QNN's results (single-crop error rate) compare with modern methods on ImageNet-$1$K Dataset.}\label{table:5}
\end{table}

As far as we known, most previous works of automatic network generation did not report competitive result on large scale image classification datasets. With the conception of block learning, we can transfer our architecture learned in small datasets to big dataset like ImageNet task easily. In the future experiments, we will try to apply the generated blocks in other tasks such as object detection and semantic segmentation.

\section{ Conclusion}

In this paper, we show how to efficiently design high performance network blocks with Q-learning. We use a distributed asynchronous Q-learning framework and an early stop strategy focusing on fast block structures searching. We applied the framework to automatic block generation for constructing good convolutional network. Our Block-QNN networks outperform modern hand-crafted networks as well as other auto-generated networks in image classification tasks. The best block structure which achieves a state-of-the-art performance on CIFAR can be transfer to the large-scale dataset ImageNet easily, and also yield a competitive performance compared with best hand-crafted networks. We show that searching with the block design strategy can get more elegant and model explicable network architectures. In the future, we will continue to improve the proposed framework from different aspects, such as using more powerful convolution layers and making the searching process faster. We will also try to search blocks with limited FLOPs and conduct experiments on other tasks such as detection or segmentation.

\section*{Acknowledgments}
This work has been supported by the National Natural Science Foundation of China (NSFC) Grants 61721004 and 61633021. 

\appendix
\section*{Appendix}
\section{Efficiency of BlockQNN}

We demonstrate the effectiveness of our proposed BlockQNN on network architecture generation on the CIFAR-$100$ dataset as compared to random search given an equivalent amount of training iterations, \ie~number of sampled networks. We define the effectiveness of a network architecture auto-generation algorithm as the increase in top auto-generated network performance from the initial random exploration to exploitation, since we aim to getting optimal auto-generated network instead of promoting the average performance.

\begin{figure}[t!]
	\centering
	\includegraphics[width=\linewidth]{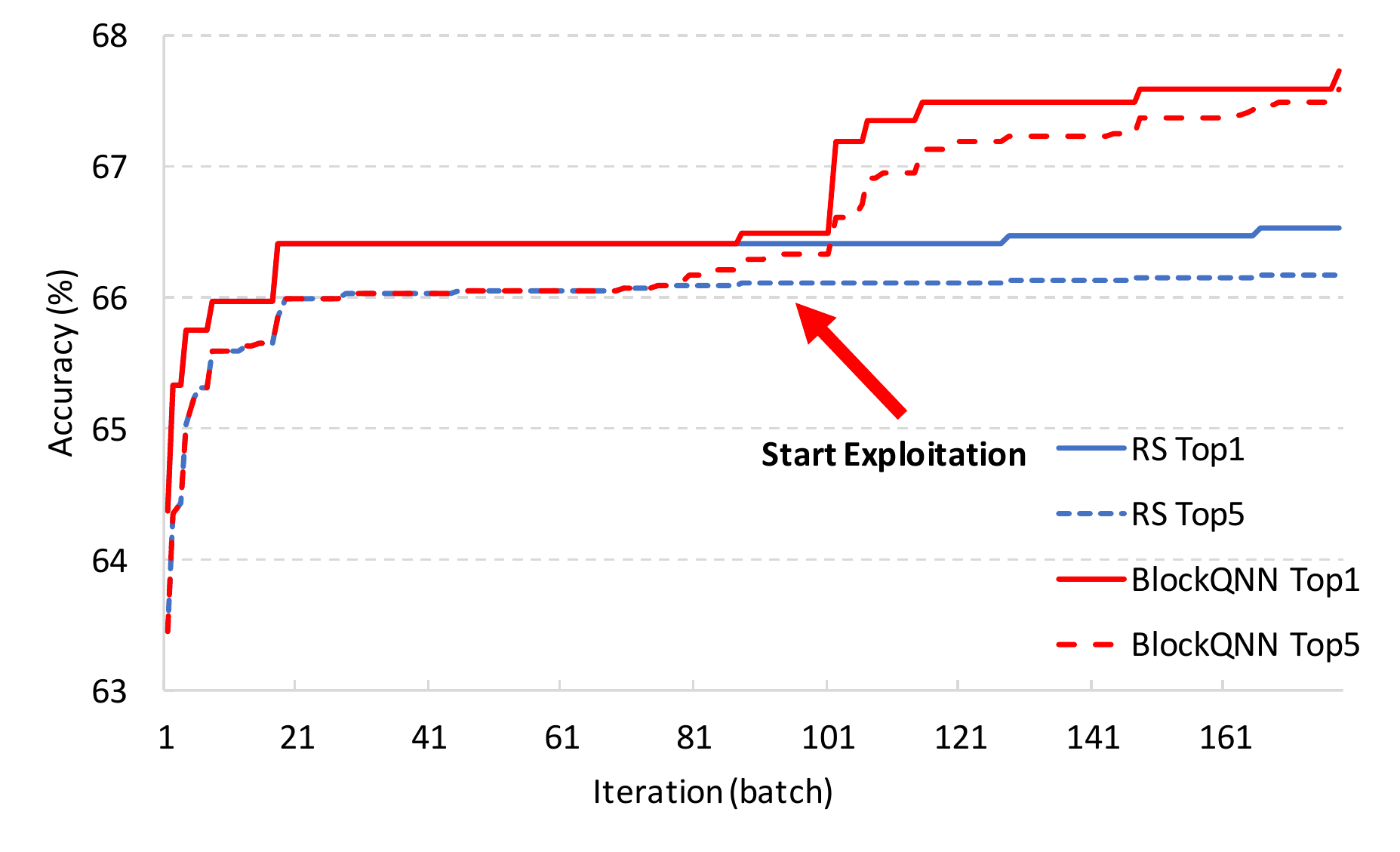}
	\caption{Measuring the efficiency of BlockQNN to random search (RS) for learning neural architectures. The x-axis measures the training iterations (batch size is $64$), \ie~total number of architectures sampled, and the y-axis is the early stop performance after $12$ epochs on CIFAR-$100$ training. Each pair of curves measures the mean accuracy across top ranking models generated by each algorithm. Best viewed in color.}
	\label{fig:1}
	\vspace{-0.1cm}
\end{figure}

Figure~\ref{fig:1} shows the performance of BlockQNN and random search (RS) for a complete training process, \ie~sampling $11, 392$ blocks in total. We can find that the best model generated by BlockQNN is markedly better than the best model found by RS by over $1\%$ in the exploitation phase on CIFAR-$100$ dataset. We observe this in the mean performance of the top-$5$ models generated by BlockQNN compares to RS. Note that the compared random search method start from the same exploration phase as BlockQNN for fairness.

\begin{figure}[t!]
	\centering
	\includegraphics[width=\linewidth]{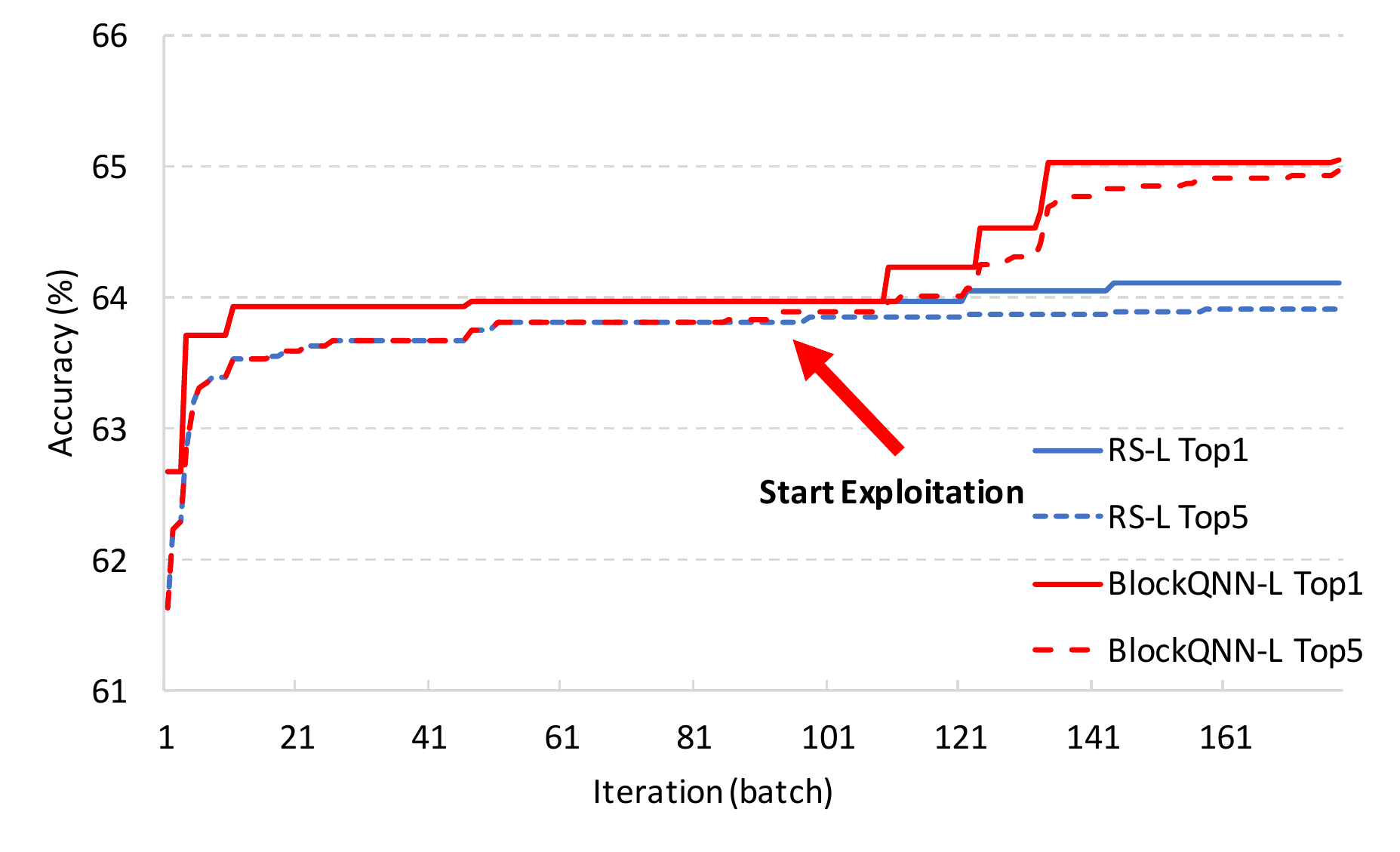}
	\caption{Measuring the efficiency of BlockQNN with limited parameters and adaptive block numbers (BlockQNN-L) to random search with limited parameters and adaptive block numbers (RS-L) for learning neural architectures. The x-axis measures the training iterations (batch size is $64$), \ie~total number of architectures sampled, and the y-axis is the early stop performance after $12$ epochs on CIFAR-$100$ training. Each pair of curves measures the mean accuracy across top ranking models generated by each algorithm. Best viewed in color.}
	\label{fig:2}
	\vspace{-0.3cm}
\end{figure}

Figure~\ref{fig:2} shows the performance of BlockQNN with limited parameters and adaptive block numbers (BlockQNN-L) and random search with limited parameters and adaptive block numbers (RS-L) for a complete training process. We can see the same phenomenon, BlockQNN-L outperform RS-L by over $1\%$ in the exploitation phase. These results prove that our BlockQNN can learn to generate better network architectures rather than random search.

\section{Evolutionary Process of Auto-Generated Blocks}

\begin{figure*}[tbp]
	\centering
	\includegraphics[width=\linewidth]{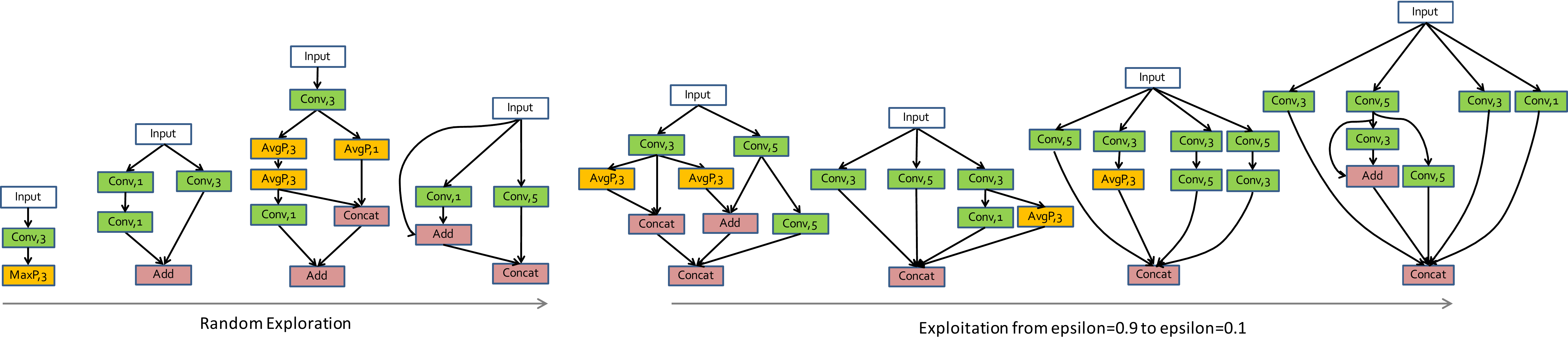}
	\caption{Evolutionary process of blocks generated by BlockQNN. We sample the block structures with median performance at iteration $[1, 30, 60, 90, 110, 130, 150, 170]$ to compare the difference between the blocks in the random exploration stage and the blocks in the exploitation stage.}
	\label{fig:3}
\end{figure*}

\begin{figure*}[tbp]
	\centering
	\includegraphics[width=\linewidth]{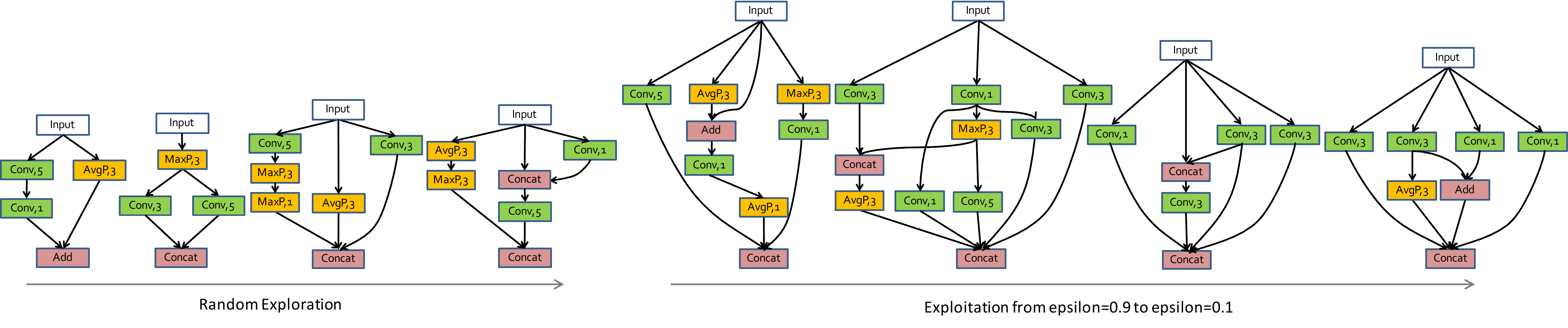}
	\caption{Evolutionary process of blocks generated by BlockQNN with limited parameters and adaptive block numbers (BlockQNN-L). We sample the block structures with median performance at iteration $[1, 30, 60, 90, 110, 130, 150, 170]$ to compare the difference between the blocks in the random exploration stage and the blocks in the exploitation stage.}
	\label{fig:4}
\end{figure*}

We sample the block structures with median performance generated by our approach in different stage, \ie~at iteration $[1, 30, 60, 90, 110, 130, 150, 170]$, to show the evolutionary process. As illustrated in Figure~\ref{fig:3} and Figure~\ref{fig:4}, \ie~BlockQNN and BlockQNN-L respectively, the block structures generated in the random exploration stage is much simpler than the structures generated in the exploitation stage.

In the exploitation stage, the multi-branch structures appear frequently. Note that the connection numbers is gradually increase and the block tend choose "Concat" as the last layer. And we can find that the short-cut connections and elemental add layers are common in the exploitation stage. Additionally, blocks generated by BlockQNN-L have less "Conv,$5$" layers, \ie~convolution layer with kernel size of $5$, since the limitation of the parameters.

These prove that our approach can learn the universal design concepts for good network blocks. Compare to other automatic network architecture design methods, our generated networks are more elegant and model explicable.

%------------------------------------------------------------------------

\section{Additional Experiment}
We also use BlockQNN to generate optimal model on person key-points task. The training process is conducted on MPII dataset, and then, we transfer the best model found in MPII to COCO challenge. It costs $5$ days to complete the searching process. The auto-generated network for key-points task outperform the state-of-the-art hourglass $2$ stacks network, \ie~$70.5$ AP compares to $70.1$ AP on COCO validation dataset.

{\small
\bibliographystyle{ieee}
\bibliography{egbib}
}

\end{document}